\documentclass[11pt]{article}

\usepackage[final]{acl}

 \usepackage{microtype}

\usepackage[english,bidi=default]{babel} 
\babelfont{rm}{TeXGyreTermesX} 
\babelprovide[import]{hindi}
\babelfont[*devanagari]{rm}{Lohit Devanagari}
\babelprovide[import]{arabic}
\babelfont[*arabic]{rm}{Noto Sans Arabic}

\usepackage{amsmath}
\usepackage{multirow}
\usepackage{booktabs}
\usepackage{graphicx}
\usepackage{algorithm}
\usepackage{algpseudocode}
\usepackage{amssymb}

\title{AFA-LoRA: Enabling Non-Linear Adaptations in LoRA with Activation Function Annealing}

\author{
    Jiacheng Li\textsuperscript{\rm 1}\thanks{\ \ These authors contributed equally to this work.},
    \ Jianchao Tan\textsuperscript{\rm 1}\footnotemark[1],
    \ Zhidong Yang\textsuperscript{\rm 2},
    \ Feiye Huo\textsuperscript{\rm 1},\\
    \ {\bf Yerui Sun}\textsuperscript{\rm 1},
    \ {\bf Yuchen Xie}\textsuperscript{\rm 1},
    \ {\bf Xunliang Cai}\textsuperscript{\rm 1}, \\
    \textsuperscript{\rm 1}Meituan, Beijing, China \\
    \textsuperscript{\rm 2}Hong Kong University of Science and Technology, Hong Kong SAR, China \\
    \texttt{lijiacheng14@meituan.com}
}

\begin{document}

\maketitle

\begin{abstract}
Low-Rank Adaptation (LoRA) is a widely adopted parameter-efficient fine-tuning (PEFT) method. However, its linear adaptation process limits its expressive power. This means there is a gap between the expressive power of linear training and non-linear training. To bridge this gap, we propose AFA-LoRA, a novel training strategy that brings non-linear expressivity to LoRA while maintaining its seamless mergeability. Our key innovation is an annealed activation function that transitions from a non-linear to a linear transformation during training, allowing the adapter to initially adopt stronger representational capabilities before converging to a mergeable linear form. We implement our method on supervised fine-tuning, reinforcement learning, and speculative decoding. The results show that AFA-LoRA reduces the performance gap between LoRA and full-parameter training. This work enables a more powerful and practical paradigm of parameter-efficient adaptation.
\end{abstract}

\section{Introduction}
\label{sec:introduction}
The growth of Large Language Models (LLMs)\cite{vaswani2017attention, team2025longcat, achiam2023gpt, liu2024deepseek, bai2023qwen} has revolutionized natural language processing. However, due to the extremely large number of parameters in these massive models, fully fine-tuning them for downstream tasks is usually infeasible. This challenge led to the development of the Parameter Efficient Fine-Tuning (PEFT) method, which aims to achieve competitive performance by training only a small fraction of the model's parameters. Among these, Low-Rank Adaptation (LoRA) \cite{hu2022lora} has become a major method. LoRA freezes the weights $W$ of the pre-trained model and inserts a trainable rank decomposition matrix (adapter) into each layer. The update is parameterized as $\Delta W = \frac{\alpha}{r} B A$, where $A \in \mathbb{R}^{r \times d_{\text{in}}}$ and $B \in \mathbb{R}^{d_{\text{out}} \times r}$ are the low-rank matrices with rank $r$, and $\alpha$ is a scaling hyperparameter. This design allows the adapter to be seamlessly merged back into the main model after training. ($W_{\text{new}} = W + \frac{\alpha}{r} BA$).

Despite its wide application, LoRA's expressive power is limited. From a design perspective, LoRA's forward propagation process is linear and lacks the nonlinear transformation capability inherent in the feedforward layer of the basic model, which is fully utilized during full parameter fine-tuning. By introducing non-linear functions into LoRA's training, we aim to reduce the difference in performance between LoRA and full fine-tuning.
A seemingly straightforward solution would be to introduce non-linear activation functions (e.g., ReLU) between the LoRA matrices. However, this approach creates a new problem: the resulting non-linear adapter can no longer be merged into the main model through simple matrix addition. 

To resolve this conflict, we propose AFA-LoRA (Activation Function Annealing LoRA), a novel training strategy that combines the advantages of nonlinear training and linear integration. Our main point is that the need for nonlinearity is especially critical in the initial training phase of the model, while the fusion capability only needs to be guaranteed to be linear at the end of training. AFA-LoRA introduces an annealed activation function, $y = \beta \cdot \sigma(x) + (1 - \beta) \cdot x$, placed between the $A$ and $B$ matrices of LoRA. The weight $\beta$ is annealed from 1 to 0 over the training process. Initially, the adapter behaves as a powerful non-linear projector ($\beta=1$), maximizing learning capacity. As training progresses, it smoothly and differentiably transitions into a linear function ($\beta=0$), guaranteeing mergeability upon convergence.

We evaluated AFA-LoRA on a variety of tasks. In the Supervised Fine-Tuning (SFT) benchmark, it narrowed the performance gap between standard LoRA and fully parametric fine-tuning. Secondly, we integrated AFA-LoRA into the GRPO framework\cite{shao2024deepseekmath} for reinforcement learning, demonstrating that it also effectively reduces the gap between GRPO-LoRA and full-parameter GRPO, showing that it works well for more than just SFT. We also integrated AFA-LoRA into the draft model in Eagle, a popular speculative decoding framework. We added LoRA adapters to the draft model in Eagle and trained them together with the main weights. This enabled the draft model to accept longer token sequences, showing that AFA-LoRA can be well adapted to different tasks.

In summary, our contributions are:
\begin{itemize}
    \item \textbf{Problem Formulation:} We point out the trade-off in LoRA between learning ability and integratability. Adding non-linearity helps models learn better, but it usually makes merging harder—a challenge for any mergeable PEFT method.
    
    \item \textbf{Method Innovation:} We introduce Activation Function Annealing (AFA), a training method that implements non-linear functions initially, then smoothly switches to linear ones for inference. This way, AFA-LoRA maintains full mergeability while boosting performance.
    
    \item \textbf{Experimental Validation:} We show that AFA-LoRA works well on supervised fine-tuning, reinforcement learning (GRPO), and speculative decoding (Eagle). In all cases, it narrows the gap with full fine-tuning and still allows for easy merging after training.
\end{itemize}

\section{Theory}
\label{sec:theory}

This section presents the theoretical background for Activation Function Annealing (AFA). We define AFA as a method that enables the exploration of both linear and non-linear function spaces during training. In addition, we analyze its advantages from an optimization perspective, showing how AFA can improve model adaptation and convergence.
\subsection{Formalization}

The fine-tuning process can be viewed as learning a parameterized function $F_{\theta}(x)$ that adapts a pre-trained model. Within this framework, an adapter module (e.g., a LoRA branch) constitutes a specific functional component. The standard LoRA adapter applies a purely linear transformation: $F_{\text{LoRA}}(x) = W_2 W_1 x$. This means there is no non-linear activation between the two matrices; it is equivalent to applying the identity function.

The core of our method is the introduction of a time-dependent activation function $\sigma_{\text{AFA}}$, defined as
\begin{equation}
\label{eq:afa}
\sigma_{\text{AFA}}(x; t) = \beta(t) \cdot \sigma(x) + (1 - \beta(t)) \cdot x,
\end{equation}
where $t \in [0, T]$ denotes the training step, and $\beta(t)$ is an annealing coefficient that decreases monotonically from $\beta(0)=1$ to $\beta(T)=0$. The function $\sigma$ is a standard non-linear activation function, such as ReLU. The resulting adapter using AFA is given by $F_{\text{AFA}}(x; t) = W_2(t) \cdot \sigma_{\text{AFA}}(W_1(t) x; t)$.

\subsection{Key Properties}

The AFA method has several features that help explain why it works well in practice. First, it satisfies clear boundary conditions: at the start of training ($t=0$), $\sigma_{\text{AFA}}(x; 0) = \sigma(x)$, endowing the adapter with full non-linear capacity; at convergence ($t=T$), $\sigma_{\text{AFA}}(x; T) = x$, reducing the adapter to a linear function that can be seamlessly merged into the main model. Second, provided $\sigma(x)$ and $\beta(t)$ are continuous, $\sigma_{\text{AFA}}(x; t)$ is continuous in its arguments, ensuring a smooth and stable optimization trajectory.

Most importantly, AFA enables a dynamic expansion of the searching space. Let $\mathcal{F}_{\text{Linear}}$ represent the space of linear adapters and $\mathcal{F}_{\text{Nonlinear}}$ the space of non-linear adapters with a fixed $\sigma$. The AFA strategy defines a continuous family of intermediate spaces $\mathcal{F}_{\text{AFA}}(t)$. This family originates from the non-linear space, $\mathcal{F}_{\text{AFA}}(0) = \mathcal{F}_{\text{Nonlinear}}$, and terminates in the linear space, $\mathcal{F}_{\text{AFA}}(T) = \mathcal{F}_{\text{Linear}}$. Crucially, for any $t < T$, the linear space is a proper subset, $\mathcal{F}_{\text{Linear}} \subset \mathcal{F}_{\text{AFA}}(t)$. This guaranties that AFA searches a strictly richer space than standard LoRA throughout most of the training process, while finally converging to a mergeable solution.

\subsection{Optimization Landscape Perspective}

The advantage of AFA can be further understood through the lens of optimization. Full fine-tuning operates on a complex, high-dimensional loss landscape $\mathcal{L}_{\text{Full}}(\Theta)$. In contrast, LoRA constrains the optimization to a lower-dimensional subspace $\mathcal{L}_{\text{LoRA}}(\theta)$, which may lack access to high quality minima present in the full landscape.

The AFA strategy can be viewed as a guided search. Initially, with $\beta \approx 1$, the optimization occurs in an expanded space $\mathcal{L}_{\text{AFA}}$, leading to the discovery of more complex and deeper feature adaptations. As $\beta$ anneals to zero, the search space keeps getting smaller, guiding the optimization trajectory from the promising region found in the expanded space back into the constrained linear subspace $\mathcal{L}_{\text{LoRA}}$. This process effectively guides the model parameters to a superior solution that can still be merged into the main model, thereby reducing the performance gap between LoRA and full fine-tuning.

\subsection{Generality of the Framework}

The AFA formulation presented in Eq.~\eqref{eq:afa} is not limited to the LoRA framework. It can be used in many cases where we want a neural network component to learn with more flexibility during training but need it to fit a certain structure at deployment. The concept can be generalized to any scenario where a target component $C_{\text{target}}(x)$ is augmented with a more expressive component $C_{\text{boost}}(x)$ during training via annealing. Using AFA in LoRA, where the final goal is a linear adapter, is a strong example of this general method.

\section{Related Work}
\label{sec:related_work}

We build on developments in PEFT and explore new adaptations of activation functions for neural networks. Here, we summarize representative related works that set the stage for our contribution.

\subsection{Parameter-Efficient Fine-Tuning}

The high cost of full fine-tuning has led to the creation of PEFT-based approaches. Early approaches include adapter-based methods, which insert small, trainable modules between layers of a pre-trained model~\cite{houlsby2019parameter}, and prompt-based techniques like prompt-tuning and prefix-tuning, which optimize continuous input vectors~\cite{lester2021power, li2021prefix}. Among these methods, Low-Rank Adaptation (LoRA)~\cite{hu2022lora} is widely used because it is efficient, and its adapters can be easily merged into the main model after training; thus, there is no extra cost during inference. In our method, we use LoRA as a starting point and focus on improving its ability to learn complex patterns while retaining the integrable features.

\subsection{Advances in the LoRA Framework}

The success of LoRA has inspired extensive research aimed at improving its efficiency and performance. One approach is adaptive parameter allocation (e.g. AdaLoRA)~\cite{zhang2023adalora}. AdaLoRA changes the rank of LoRA matrices during training to better utilize computational resources. Other innovative methods include VeRA~\cite{kopiczko2023vera}, which focuses on parameter reduction, and DoRA~\cite{liu2024dora}, which decouples the magnitude and direction of weight updates. QLoRA~\cite{dettmers2023qlora} allows for fine-tuning large models using quantized weights, which saves memory and speeds up training.

Most of these methods explore how to set up or train the LoRA, but the core idea of linear adaptation remains unchanged. To address this issue, some recent methods attempt to change the internal structure of adapters. Among the aforementioned works, our method stands out because it directly tackles the trade-off between learning ability and easy merging. As far as we know, AFA-LoRA is the first to add a time-dependent non-linearity during training that gradually fades away, allowing the adapter to learn more at first and then merge smoothly into the main model.

\subsection{Activation Functions in Model Adaptation}

Activation functions can effectively introduce non-linearity into neural networks. They have been widely used in pre-training and full fine-tuning, but the exploration of their applications within PEFT adapters is still limited. Most PEFT adapters only retain the activations within the main model and do not add new non-linear components to their adapters. In our method, we propose introducing a temporary non-linear activation to the adapter during training, which helps to improve performance. This idea is similar to activation annealing used elsewhere, but here we apply it in PEFT while making sure that merging after training remains easy.

Previous work such as PReLU~\cite{he2015delving} has shown that allowing the activation function to change during training is feasible. Our method follows this idea by gradually changing the non-linearity in adapters over time.

\subsection{Summary and Positioning}

Our proposed AFA-LoRA stands out among PEFT methods. Instead of just changing LoRA's parameters, it improves the performance of adapters by introducing a temporary non-linear component during training. This non-linearity will converge to linear space complexity over the iterations, so there is no extra cost when using the model for inference. Our method directly addresses the trade-off between learning ability and integratability, providing a flexible solution that can help build better integrable adapters.

\section{Method}
\label{sec:method}

In this part, we will present details about our proposed AFA-LoRA method. To better understand the motivation behind AFA-LoRA, we will recap the standard LoRA as a preliminary first.

\subsection{Preliminaries: Low-Rank Adaptation (LoRA)}

LoRA approximates the weight update of a pre-trained matrix $W_0 \in \mathbb{R}^{d_{\text{out}} \times d_{\text{in}}}$ with a low-rank decomposition. The forward pass can be formulated as follows:
\begin{equation}
    h = W_0 x + \Delta W x = W_0 x + B A x,
\end{equation}
where $A \in \mathbb{R}^{r \times d_{\text{in}}}$, $B \in \mathbb{R}^{d_{\text{out}} \times r}$ are trainable matrices with rank $r \ll \min(d_{\text{in}}, d_{\text{out}})$. After training with LoRA, the forward pass is merged as $W' = W_0 + BA$, resulting in zero inference overhead. However, the linearity of $\Delta W$ does not fully explore the expressive power of the pre-trained main model.

\subsection{Activation Function Annealing (AFA)}

To enhance expressivity while preserving integrability, we insert an annealed activation function between $A$ and $B$. The adapter starts as a non-linear function and converges to a linear function.

\subsubsection{Annealed Activation Function}
We set up the annealed activation function $\phi$ by smoothly mixing a non-linear function $\sigma$ (e.g., ReLU) with the identity function:
\begin{equation}
\label{eq:afa_core}
    \phi(x; \beta) = \beta \cdot \sigma(x) + (1 - \beta) \cdot x,
\end{equation}

Where, $\beta$ is a value that decreases from 1 to 0 as training continues. This means that when $\beta=1$, we will adapted non-linear function $\sigma(x)$ for updating, and when $\beta=0$, we simply use $x$.
\subsubsection{AFA-LoRA Forward Computation}

\begin{figure}[h!]
\centering
\includegraphics[width=0.8\linewidth]{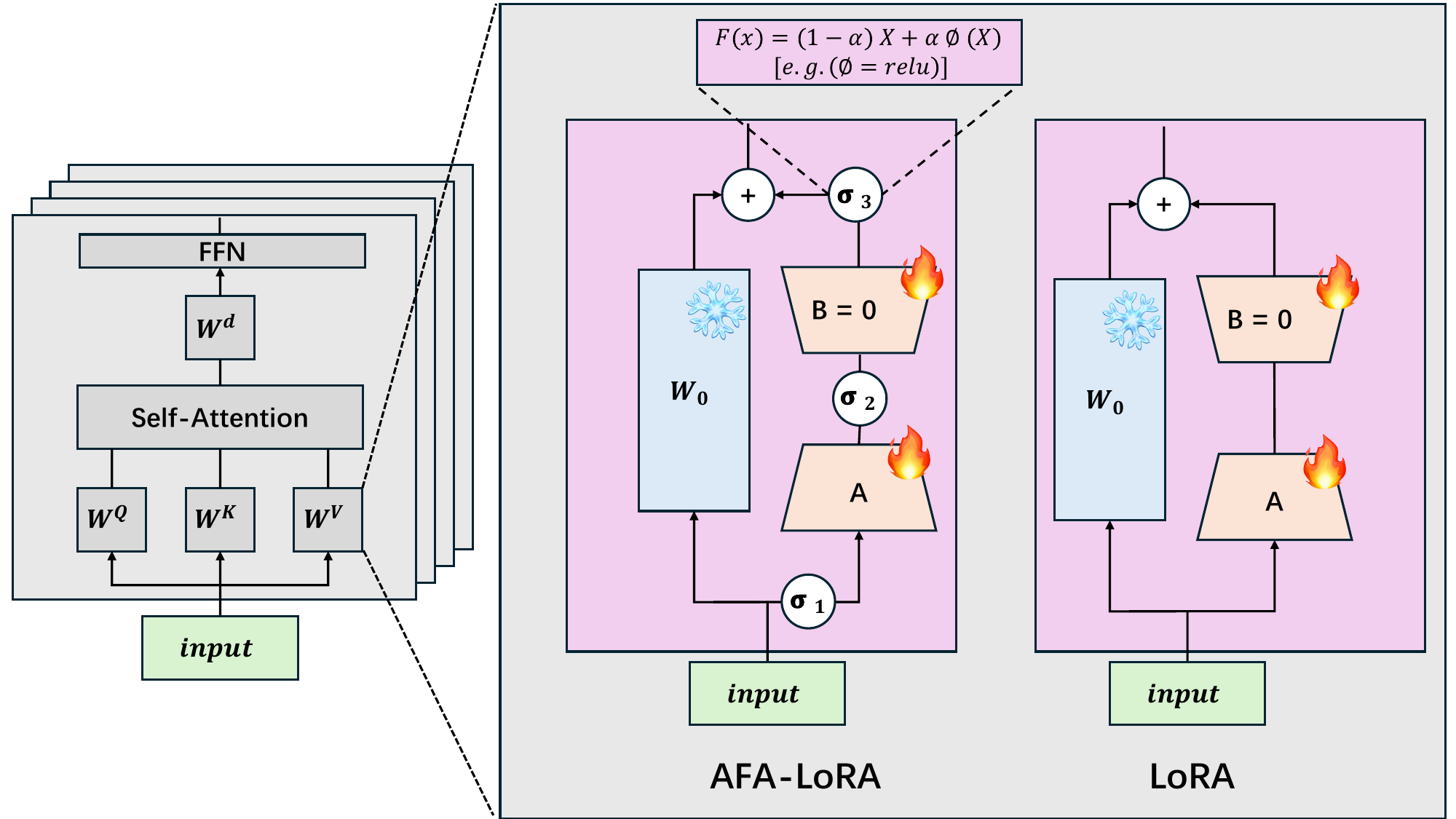} 
\caption{Architectural comparison of (a) Standard LoRA and (b) AFA-LoRA. The annealed activation function $\phi$ is placed beside the $A$ and $B$ matrices.}
\label{fig:afa_arch}
\vspace{-1em}
\end{figure}

The AFA-LoRA forward pass is given by:
\begin{equation}
\label{eq:afa_lora_forward}
\begin{split}
    h &= W_0 x + B \phi(A x; \beta(t)) \\
      &= W_0 x + B \left[ \beta(t) \sigma(A x) + (1 - \beta(t)) A x \right]
\end{split}
\end{equation}
Where, $t$ is the training step. The architectural change is shown in Figure~\ref{fig:afa_arch}. At the end of training ($\beta(T)=0$), this calculation will converge to $h = W_0 x + B A x$, which means all the extra weights can be easily merged into the main model.

\subsection{Annealing Schedule}

\begin{figure}[h!]
\centering
\includegraphics[width=0.8\linewidth]{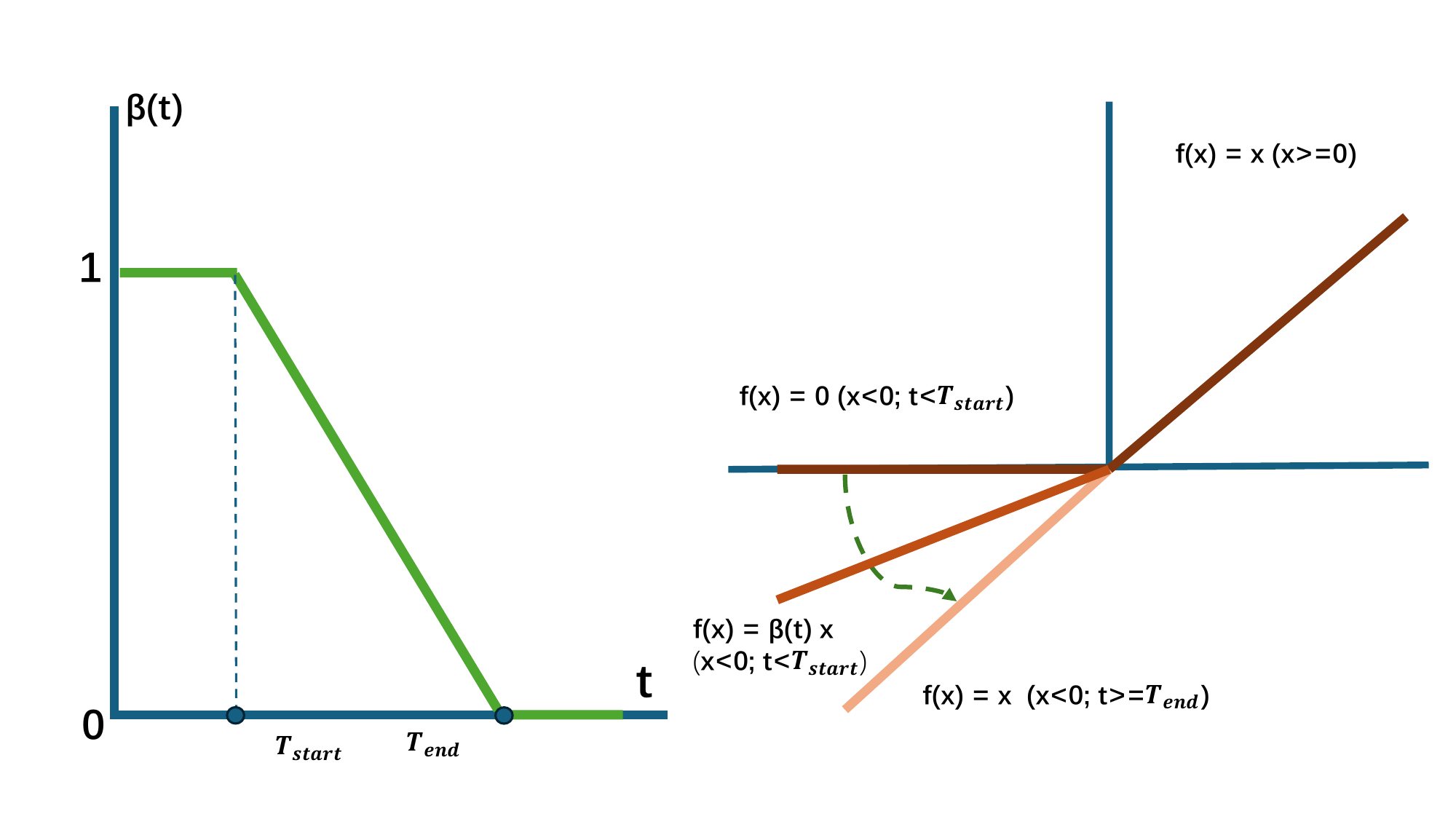} 
\caption{Illustration of ReLU-to-linear activation annealing and the decay schedule for $\beta(t)$ during training.}
\label{fig:beta_schedule}
\vspace{-1em}
\end{figure}

The annealing schedule for $\beta(t)$ is defined over a range of training steps. A linear schedule is formulated as:
\begin{equation}
\label{eq:beta_schedule}
    \beta(t) = \max\left(0, \: 1 - \frac{\max(0, t - T_{\text{start}})}{T_{\text{end}} - T_{\text{start}}} \right).
\end{equation}
In the experiment, unless otherwise specified, setting $T_{\text{start}}=0$ and $T_{\text{end}} = 0.3T$ anneals $\beta$ over the first 30\% of training. The schedule profile is illustrated in Figure~\ref{fig:beta_schedule}.

\subsection{Training Algorithm}

Algorithm~\ref{alg:afa_lora} summarizes the training steps for AFA-LoRA. The proposed AFA-LoRA is an effective extension of standard LoRA by only adapting a time scheduler $\beta(t)$ and non-linear function $\phi$ during each forward pass.
\begin{algorithm}[h!]
    \caption{AFA-LoRA Training}
    \label{alg:afa_lora}
    \begin{algorithmic}[1] 
        \State \textbf{Input:} Model $W_0$, Dataset $D$, Total steps $T$, Annealing range $[T_{\text{start}}, T_{\text{end}}]$
        \State Initialize LoRA parameters $\theta = (A, B)$.
        \For{$t=1$ to $T$ }
            \State $\beta(t) = 1 - (t-T_\mathrm{{start}})/(T_\mathrm{{end}}-T_\mathrm{{start}})$
            \State Sample batch $(x, y)~{\sim}~D$
            \State Forward pass: $h = W_0 x + B[\beta(t)\sigma(Ax)+(1-\beta(t))Ax]$
            \State Compute loss $\mathcal L$
            \State Update $\theta$ via gradient descent
        \EndFor
        \State Merge adapter: $W' = W_0 + BA$
    \end{algorithmic}
\end{algorithm}

\section{Experiments}
\label{sec:experiments}
We applied our AFA-LoRA across three scenarios to comprehensively evaluate its performance. In the supervised fine-tuning domain, we evaluate the method's capability to enhance commonsense reasoning using the Llama-3-8B model on the Commonsense-170K dataset, with performance evaluated across eight diverse benchmarks, including ARC-Challenge, BoolQ, and HellaSwag, through accuracy metrics. In the experiment of reinforcement learning, we integrate AFA-LoRA into the GRPO framework to optimize mathematical problem-solving policies on the GSM8K dataset, employing Qwen2.5 models ranging from 3B to 32B parameters and quantifying improvements through reward gains and reductions in performance gaps relative to full fine-tuning. Finally, in speculative decoding, we jointly train the draft model and AFA-LoRA adapters within the Eagle framework on the ShareGPT dataset using Llama3.1-8B and evaluate token acceptance rates. Each experimental paradigm employs distinct model architectures, adaptation strategies, and evaluation methodologies to validate AFA-LoRA's capabilities across diverse scenarios.

\subsection{Supervised Fine-Tuning Experiments}
\label{subsec:sft_experiments}

\begin{table*}[h!]
\caption{Commonsense reasoning evaluation results (Accuracy \%) on Llama-3-8B. Results show the 30\% decay configuration for each AFA placement variant. Avg is the macro-average across all 8 tasks. The best result for each method is \textbf{bold}.}
\label{tab:main_results}
\centering
\footnotesize
\setlength{\tabcolsep}{2pt}
\renewcommand{\arraystretch}{0.85}
\begin{tabular}{@{}l l c c c c c c c c c@{}}
\toprule
\textbf{Method} & \textbf{Placement} & \textbf{ARC-C} & \textbf{ARC-E} & \textbf{BoolQ} & \textbf{HellaSwag} & \textbf{OpenBookQA} & \textbf{PIQA} & \textbf{Social IQA} & \textbf{WinoGrande} & \textbf{Avg} \\
\midrule
FULL-SFT & -- & 82.94 & 92.68 & 74.98 & 96.71 & 89.40 & 90.26 & 83.21 & 86.11 & 87.07 \\
\midrule
LoRA Baseline & -- & 80.03 & 90.99 & 74.56 & 96.03 & 87.00 & 88.63 & 81.17 & 86.11 & 85.57 \\
\midrule
\multirow{7}{*}{AFA-LoRA} 
 & $\sigma$-A-B & 81.23 & 91.41 & 75.57 & 95.72 & 86.00 & 88.08 & 81.93 & 85.95 & 85.74 \\
 & A-$\sigma$-B & 79.78 & 91.25 & 75.60 & 96.03 & 86.80 & 88.52 & 81.37 & 86.66 & 85.75 \\
 & A-B-$\sigma$ & 81.66 & 91.46 & 74.80 & 95.66 & 87.20 & 89.06 & 81.99 & 86.42 & \underline{86.03} \\
 & $\sigma$-A-$\sigma$-B & 80.38 & 90.91 & 74.95 & 96.11 & 87.40 & 88.52 & 80.04 & 87.45 & 85.72 \\
 & A-$\sigma$-B-$\sigma$ & 80.72 & 91.41 & 75.08 & 95.87 & 86.40 & 89.17 & 81.06 & 87.37 & 85.89 \\
 & $\sigma$-A-B-$\sigma$ & 81.14 & 91.33 & 75.44 & 95.71 & 88.40 & 88.68 & 81.53 & 87.06 & \textbf{86.16} \\
 & $\sigma$-A-$\sigma$-B-$\sigma$ & 80.29 & 90.82 & 75.96 & 95.74 & 87.20 & 89.06 & 80.60 & 85.87 & 85.69 \\
\midrule
DORA Baseline & -- & 80.46 & 90.45 & 75.66 & 95.77 & 85.00 & 87.92 & 81.06 & 87.29 & 85.45 \\
\midrule
\multirow{7}{*}{AFA-DORA} 
 & $\sigma$-A-B & 80.37 & 91.07 & 75.19 & 95.23 & 85.60 & 89.11 & 81.11 & 87.21 & 85.61 \\
 & A-$\sigma$-B & 79.69 & 91.25 & 75.93 & 95.80 & 86.40 & 88.74 & 80.45 & 86.74 & 85.63 \\
 & A-B-$\sigma$ & 80.97 & 91.62 & 75.56 & 95.90 & 88.40 & 89.11 & 82.59 & 86.58 & \textbf{86.34} \\
 & $\sigma$-A-$\sigma$-B & 80.20 & 91.07 & 75.99 & 96.05 & 87.60 & 89.33 & 81.16 & 87.21 & 86.08 \\
 & A-$\sigma$-B-$\sigma$ & 79.52 & 90.74 & 75.35 & 95.75 & 85.40 & 89.88 & 81.21 & 87.05 & 85.61 \\
 & $\sigma$-A-B-$\sigma$ & 81.48 & 91.58 & 76.17 & 95.66 & 86.80 & 89.00 & 81.42 & 86.97 & \underline{86.14} \\
 & $\sigma$-A-$\sigma$-B-$\sigma$ & 81.65 & 91.07 & 76.60 & 95.70 & 87.00 & 88.57 & 81.26 & 87.21 & 86.13 \\
\bottomrule
\end{tabular}
\end{table*}

We include both LoRA and DoRA (Weight-Decomposed Low-Rank Adaptation)~\cite{liu2024dora} as baselines in our experiments. DoRA is an advanced parameter-efficient fine-tuning method that splits weight updates into direction and magnitude, enabling it to capture more complex changes than standard LoRA.

We implemented activation function annealing with LoRA and DoRA on supervised fine-tuning with the Llama-3-8B~\cite{llama3modelcard} model and the Commonsense-170K~\cite{hu2023llm} dataset, which is designed for commonsense reasoning. We evaluate performance across eight benchmarks: ARC-Challenge~\cite{allenai:arc}, ARC-Easy, BoolQ~\cite{clark2019boolq}, HellaSwag~\cite{zellers2019hellaswag}, OpenBookQA~\cite{OpenBookQA2018}, PIQA~\cite{Bisk2020}, Social IQA~\cite{sap2019socialiqa}, and WinoGrande~\cite{sakaguchi2021winogrande}. Our experiments compare AFA-enhanced methods with three baselines: full parameter fine-tuning (Full-SFT), standard LoRA, and DoRA. For both LoRA and DoRA, we try seven different ways of placing the annealed activation function in the adapter structure, using rank $r=32$ and scaling factor $\alpha=64$. The value of $\beta$ decreases linearly over the first 30\% of training steps, so in Equation~\eqref{eq:beta_schedule}, we set $T_{start}=0$ and $T_{end}=\frac{T}{3}$.

Table~\ref{tab:main_results} shows that both AFA-LoRA and AFA-DoRA provide clear improvements over their baselines. The best AFA-LoRA setup achieves an average accuracy of 86.16\%, which is 0.59\% higher than the standard LoRA (85.57\%). For DoRA, the top AFA-DoRA variant reaches 86.34\%, a gain of 0.89\% over the DoRA baseline (85.45\%). Importantly, the best AFA-DoRA version reduces the gap to full fine-tuning (87.07\%) by about 54.94\%, while the best AFA-LoRA reduces this gap by around 39.33\%.

In conclusion, our experimental results demonstrate that using activation function annealing with LoRA or DoRA improve accuracy on commonsense reasoning tasks. This approach contributes to narrow the difference between lightweight tuning methods and full fine-tuning technique.

\subsection{Reinforcement Learning Experiments}
\label{subsec:rl_experiments}

\begin{table*}[h!]
\caption{Comprehensive GRPO Evaluation on GSM8K: Training and Validation Performance}
\label{tab:grpo_full_results}
\centering
\setlength{\tabcolsep}{5pt}
\renewcommand{\arraystretch}{0.85}
\footnotesize
\begin{tabular}{@{}l l c c c c c c c c c c@{}}
\toprule
\textbf{Model} & \textbf{Metric} & \textbf{Full} & \textbf{LoRA} & \textbf{$\sigma$A-B} & \textbf{A$\sigma$B} & \textbf{AB$\sigma$} & \textbf{$\sigma$A$\sigma$B} & \textbf{$\sigma$AB$\sigma$} & \textbf{A$\sigma$B$\sigma$} & \textbf{$\sigma$A$\sigma$B$\sigma$} \\
\midrule
\multirow{4}{*}{\textbf{3B}} 
& Train Reward & 97.50 & 95.27 & 95.41 & 95.63 & 95.61 & 95.72 & 95.86 & \textbf{96.04} & 95.68 \\
& Val Reward & 88.32 & 87.19 & 87.26 & \textbf{88.70} & 87.34 & 88.02 & 87.87 & 88.40 & 88.17 \\
& Val Gain & - & 0.0\% & 6.2\% & \textbf{133.6\%} & 13.3\% & 73.5\% & 60.2\% & 107.1\% & 86.7\% \\
\midrule

\multirow{4}{*}{\textbf{7B}}
& Train Reward & 98.77 & 97.13 & 97.17 & \textbf{97.60} & 97.50 & 97.44 & 97.36 & 97.15 & 97.25 \\
& Val Reward & 92.80 & 92.19 & 93.03 & 92.34 & \textbf{93.10} & 92.65 & 92.87 & 92.65 & 92.65 \\
& Val Gain & - & 0.0\% & 137.7\% & 24.6\% & \textbf{149.2\%} & 75.4\% & 111.5\% & 75.4\% & 75.4\% \\
\midrule

\multirow{4}{*}{\textbf{14B}}
& Train Reward & 98.83 & 97.29 & 97.46 & \textbf{97.93} & 97.56 & 97.46 & 97.54 & 97.64 & 97.54 \\
& Val Reward & 95.45 & 94.47 & 95.22 & \textbf{95.83} & 95.15 & 95.07 & 95.75 & 95.30 & 94.84 \\
& Val Gain & - & 0.0\% & 76.5\% & \textbf{138.8\%} & 69.4\% & 61.2\% & 130.6\% & 84.7\% & 37.8\% \\
\midrule

\multirow{4}{*}{\textbf{32B}}
& Train Reward & 99.10 & 97.03 & 97.32 & 97.48 & 97.38 & 97.44 & \textbf{97.50} & 97.15 & 97.27 \\
& Val Reward & 96.66 & 96.21 & 96.44 & 96.44 & 96.51 & 96.44 & 96.21 & \textbf{96.82} & 96.66 \\
& Val Gain & - & 0.0\% & 51.1\% & 51.1\% & 66.7\% & 51.1\% & 0.0\% & \textbf{135.6\%} & 100.0\% \\
\bottomrule
\end{tabular}

\vspace{0.5em}
\begin{minipage}{\textwidth}
\footnotesize
\textit{Note:} \\
- \textbf{Gain} represents the percentage improvement over standard LoRA gap reduction: 
  $\text{Gain} = \frac{\text{Method Reward} - \text{LoRA Reward}}{\text{Full Reward} - \text{LoRA Reward}} \times 100\%$ \\
- $\sigma$ indicates the position of the ReLU activation relative to LoRA matrices A/B \\
- \textbf{Bold values} denote the best performance per metric and model size \\
- Gain values exceeding 100\% indicate the method outperformed the Full-Train baseline \\
\end{minipage}
\vspace{-2.5em}
\end{table*}

We conduct reinforcement learning experiments using the GRPO (Group Relative Policy Optimization) framework, implemented with the Verl training system. This method uses preference optimization within groups of responses to keep policy updates stable. We test Qwen2.5-Instruct models of different sizes (3B, 7B, 14B, and 32B)\cite{qwen2.5, qwen2} on GSM8K\cite{cobbe2021gsm8k}, a dataset for math reasoning tasks. For training, we use the AdamW optimizer with model-specific learning rates ($2\times10^{-5}$ for 3B; $1.5\times10^{-5}$ for 7B/14B; $1.3\times10^{-5}$ for 32B). The global batch size is set to 1024 and is split into PPO mini-batches of 256 samples for each global batch. Each sequence has up to 512 prompt tokens and up to 1024 response tokens. The loss includes KL divergence regularization ($\beta=0.001$), but no entropy term is used. For each prompt, five responses are generated using vLLM~\cite{kwon2023efficient}. We adapt gradient checkpointing and FSDP to save memory and speed up training~\cite{zhao2023pytorch}. All experiments are carried out with a total of fifteen epochs. All variants of LoRA are set with rank $r=64$ and scaling factor $\alpha=32$. For AFA-LoRA, $\beta$ decays linearly over the first $30\%$ steps in all tested placements.

Table~\ref{tab:grpo_full_results} shows that AFA-LoRA is highly versatile across different evaluation scenarios. Importantly, this method not only consistently narrows the performance gap with full parameter fine-tuning but also often surpasses it on validation metrics. This is a significant achievement for a parameter-efficient approach. For example, the AB$\sigma$ configuration achieves a 149.2\% improvement at the 7B scale, which means it generalizes even better than full fine-tuning in some cases. These consistent gains across model sizes show substantial task-agnostic benefits, especially for mid-sized models where annealed non-linearity provides significant benefits. All these improvements are achieved while maintaining full intergratability after training, so there is no extra inference cost, and deployment remains practical compared to traditional fine-tuning methods.

Overall, these results demonstrate that AFA-LoRA delivers clear and consistent improvements to reinforcement learning tasks across a wide range of model sizes. The method is especially effective for mid-sized models, where it often achieves even better generalization than full fine-tuning. These findings highlight the value of activation function annealing as a simple yet powerful technique for improving adaptation quality in large language models under RL optimization.

\subsection{Speculative Sampling Experiments}
\label{subsec:speculative_experiments}

We apply AFA-LoRA adapter to every linear layer in Eagle's draft model. Each adapter uses an activation function that gradually decays from non-linear and linear transformation during training.

For each input $x$, the output is:
\begin{equation}
    y = W_\mathrm{main}\,x + B[\;\beta\sigma(A{x})+(1-\beta)A{x}]
\end{equation}

where $A,B$ are trainable low-rank matrices, $\sigma$ is an activation function (ReLU/GeLU/SiLU), and $\beta$ decays from 1 to 0 during training. 

Joint training with AFA-LoRA adapters enhances the draft model's capacity, particularly beneficial for smaller models like Eagle. Post-training, adapters merge into base weights without inference overhead.



We evaluated our method on ShareGPT dataset with Llama-3.1-8B as the draft model, using AdamW optimizer with a learning rate of $6\times10^{-5}$ in BF16 precision.

\begin{figure}[h!]
\centering
\includegraphics[width=\linewidth]{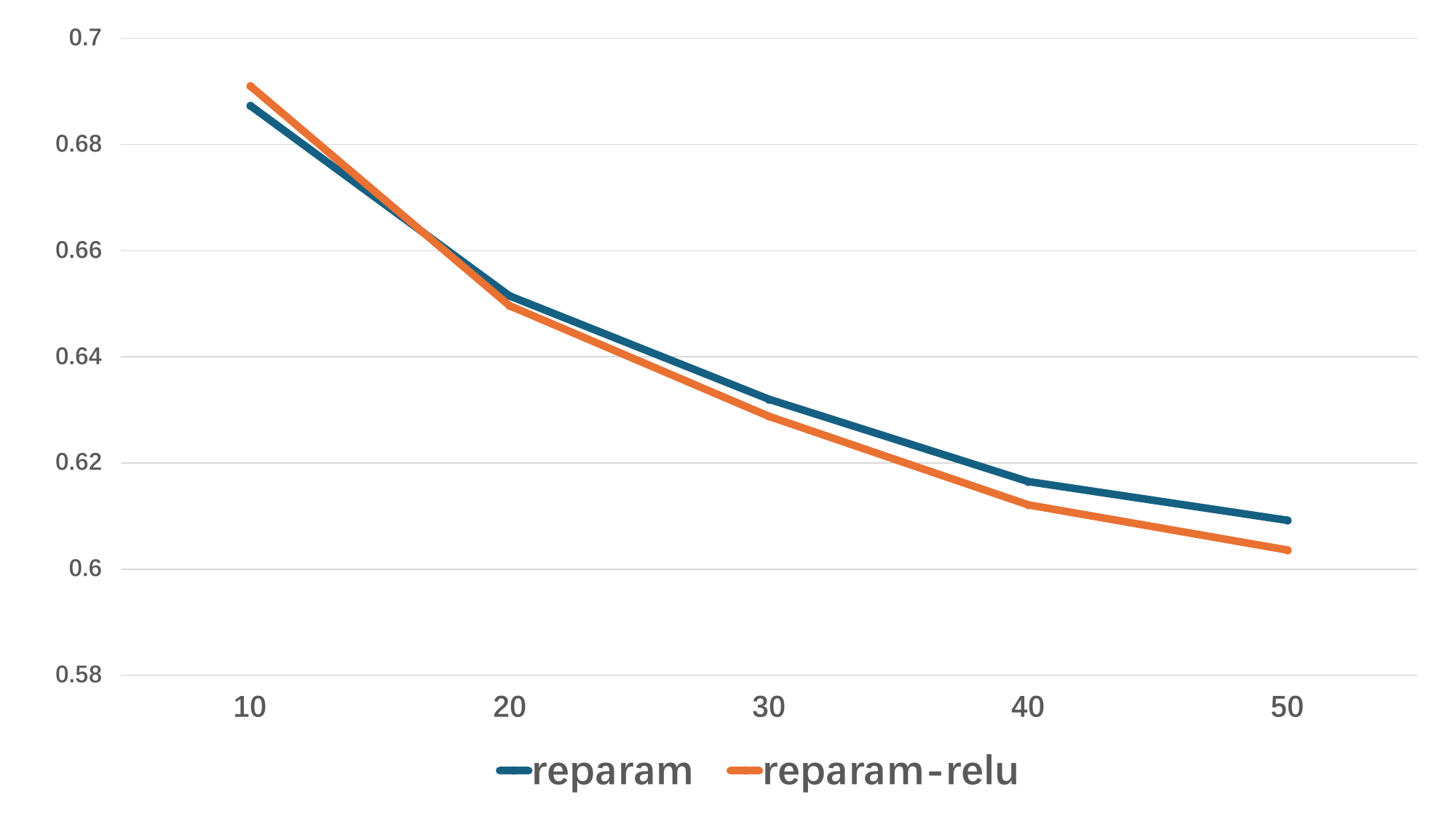} 
\caption{Llama3.1-8b training loss-epoch curves for Eagle-1 on the ShareGPT dataset}
\label{fig:eagle-line}
\vspace{-0.5em}
\end{figure}

\begin{table}[h!]
\caption{Average accepted tokens per prompt for speculative decoding on ShareGPT (Llama-3.1-8B), comparing standard Eagle to versions jointly trained with LoRA adapters and various activation annealing methods, for both chain ($d=4$) and tree ($d=5$) models.}
\label{tab:speculative_results}
\centering
\small
\setlength{\tabcolsep}{1.5pt}
\begin{tabular}{@{}lcccc@{}}
\toprule
\multicolumn{5}{c}{\textbf{Chain Decoding}} \\
\midrule
\textbf{Activation} & \textbf{MT-Bench} & \textbf{GSM8K} & \textbf{HumEval} & \textbf{Alpaca} \\
\midrule
Eagle & 1.6974 & 1.8861 & 2.2523 & 1.5810 \\
Eagle\_LoRA & 1.6931 & 1.9208 & 2.2758 & 1.6032 \\
Eagle\_LoRA\_ReLU & \underline{1.7297} & 1.9230 & 2.2711 & \underline{1.6077} \\
Eagle\_LoRA\_SiLU & \textbf{1.7306} & \underline{1.9308} & \textbf{2.2981} & \textbf{1.6349} \\
Eagle\_LoRA\_GeLU & 1.7161 & \textbf{1.9330} & \underline{2.2841} & 1.5925 \\
\midrule
\multicolumn{5}{c}{\textbf{Tree Decoding}} \\
\midrule
\textbf{Activation} & \textbf{MT-Bench} & \textbf{GSM8K} & \textbf{HumEval} & \textbf{Alpaca} \\
\midrule
Eagle & 3.1406 & 3.4137 & 3.8272 & 3.0851 \\
Eagle\_LoRA & 3.1803 & 3.4765 & 3.8609 & \textbf{3.1677} \\
Eagle\_LoRA\_ReLU & \textbf{3.2124} & 3.4479 & \underline{3.8806} & 3.1514 \\
Eagle\_LoRA\_SiLU & \underline{3.2067} & \underline{3.4849} & \textbf{3.8933} & \underline{3.1654} \\
Eagle\_LoRA\_GeLU & 3.2020 & \textbf{3.4891} & 3.8693 & 3.1559 \\
\bottomrule
\end{tabular}
\vspace{0.2em}
\parbox{\linewidth}{
\footnotesize
\textit{Note:}\\ - Metrics represent average accepted tokens per prompt. AFA-LoRA with ReLU, SiLU and GeLU use step decay (5 epoch decay and 45 epoch stay linear).
}
\end{table}

Table~\ref{tab:speculative_results} shows that applying AFA-LoRA adapters and training the draft model in Eagle with the adapters contributes to better results than using the standard Eagle alone. In both chain and tree decoding setups, Eagle\_LoRA achieves a higher average of accepted tokens per prompt compared to naive Eagle, demonstrating that this joint training approach improves the draft model's capacity for speculative decoding.

\begin{table*}[h!]
\caption{Commonsense reasoning evaluation results (Accuracy \%) with different decay steps and activation functions. Results are grouped by method (DORA/LoRA) and activation function (GeLU/SiLU). The best average for each method is \textbf{bold}.}
\vspace{-0.5em}
\label{tab:decay_results}
\centering
\footnotesize
\setlength{\tabcolsep}{2pt}
\renewcommand{\arraystretch}{0.85}
\begin{tabular}{@{}l l c c c c c c c c c c@{}}
\toprule
\textbf{Method} & \textbf{Activation} & \textbf{Decay} & \textbf{ARC-C} & \textbf{ARC-E} & \textbf{BoolQ} & \textbf{HellaSwag} & \textbf{OpenBookQA} & \textbf{PIQA} & \textbf{Social IQA} & \textbf{WinoGrande} & \textbf{Avg} \\
\midrule
\multirow{3}{*}{DORA} & \multirow{3}{*}{GeLU} & 30\% & 82.00 & 91.46 & 75.81 & 95.79 & 87.60 & 88.63 & 81.01 & 85.48 & \textbf{85.97} \\
 & & 60\% & 81.57 & 90.66 & 75.65 & 95.73 & 86.20 & 89.34 & 81.47 & 86.19 & 85.85 \\
 & & 100\% & 80.72 & 90.91 & 74.68 & 95.72 & 86.20 & 88.74 & 80.91 & 85.87 & 85.47 \\
\cmidrule(lr){2-12}
 & \multirow{3}{*}{SiLU} & 30\% & 80.20 & 91.46 & 75.26 & 95.95 & 87.40 & 88.79 & 81.06 & 85.64 & \textbf{85.72} \\
 & & 60\% & 81.31 & 91.08 & 75.54 & 95.54 & 87.60 & 88.63 & 80.45 & 85.64 & \textbf{85.72} \\
 & & 100\% & 78.75 & 90.99 & 75.35 & 95.74 & 88.20 & 88.08 & 81.27 & 86.27 & 85.58 \\
\midrule
\multirow{3}{*}{LoRA} & \multirow{3}{*}{GeLU} & 30\% & 80.80 & 91.12 & 76.42 & 95.96 & 86.60 & 89.39 & 80.45 & 86.90 & \textbf{85.96} \\
 & & 60\% & 81.48 & 91.33 & 75.87 & 95.89 & 86.80 & 88.90 & 80.45 & 86.50 & 85.90 \\
 & & 100\% & 80.03 & 90.95 & 74.71 & 95.68 & 85.40 & 89.17 & 81.27 & 86.11 & 85.41 \\
\cmidrule(lr){2-12}
 & \multirow{3}{*}{SiLU} & 30\% & 80.20 & 90.95 & 74.83 & 95.89 & 87.40 & 89.23 & 80.55 & 86.66 & \textbf{85.71} \\
 & & 60\% & 79.52 & 90.70 & 75.08 & 95.99 & 87.00 & 89.12 & 81.47 & 86.35 & 85.65 \\
 & & 100\% & 78.67 & 90.53 & 75.69 & 95.56 & 87.00 & 88.74 & 80.91 & 86.27 & 85.42 \\
\bottomrule
\end{tabular}
\end{table*}

SiLU \cite{elfwing2018sigmoid} annealing yields the highest acceptance rates (Table~\ref{tab:speculative_results}), with +1.0\%/+0.8\% gains on HumanEval for chain/tree decoding versus LoRA baseline.


Figure~\ref{fig:eagle-line} shows the progressive advantage of AFA-LoRA during training, showing the loss curves of the Llama-3.1-8B Eagle-1 models on the ShareGPT dataset. While both the pure LoRA baseline and the AFA-LoRA based on ReLU start from similar initial loss values, their trajectories rapidly diverge as training progresses. The key observation is the increasing divergence between the trajectories as the training continues. The gap of performance between naive LoRA and AFA-LoRA demonstrates that the annealed non-linearity provides compounding benefits throughout training, with its relative advantage growing more pronounced during the later optimization stages when fine-grained feature refinement becomes critical.

Overall, adding AFA-LoRA adapters contributes to performance gain of Eagle in accepting tokens during speculative decoding. This shows that our method is a strong choice for improving draft models in real-world generation tasks.

\section{Ablation Studies}

In this section, we explore the impact of different annealing schedules and activation functions on AFA-LoRA performance. While our main experiments (Section~\ref{sec:experiments}) focused on ReLU activation with a 30\% decay duration, here we evaluate SiLU and GeLU activations across three decay schedules. We performed ablation experiments using the same Llama3-8B LoRA SFT task described in Section ~\ref{subsec:sft_experiments}, keeping the same hyperparameters (rank $r=32$, scaling factor $\alpha=64$) and evaluation benchmarks. The key difference lies in the annealing configurations: instead of changing the position of activation function as in the table ~\ref{tab:main_results}, we fix the position with the manner of A-$\sigma$-B and change the activation function type ($\sigma \in \{\text{SiLU}, \text{GeLU}\}$) and the decay schedules ($T_{\text{end}} \in \{0.3T, 0.6T, T\}$).

Table~\ref{tab:decay_results} presents comprehensive results across all configurations. Across both LoRA and DoRA variants, the 30\% decay schedule consistently achieves the best average performance. For LoRA with GeLU, the 30\% decay yields 85.96\% average accuracy, outperforming 60\% decay (85.90\%) and 100\% decay (85.41\%). Similarly, DoRA with GeLU peaks at 85.97\% with 30\% decay. This pattern holds across activation functions, demonstrating that early annealing followed by extended linear training is the optimal strategy.

Figure~\ref{fig:eagle-line} shows why early annealing works. During the decay phase, models with activation annealing show slightly higher training loss than the baseline. However, as the training steps going further, the loss of our method outperforms the baseline with a more optimal convergence. This indicates that the nonlinear patterns captured during early training create a stronger foundation for later optimization. The training process reveals that the nonlinear patterns captured during the early training contribute to the performance gain and improve the optimality of convergence in the later training steps. 

These ablation studies confirm our key insight: AFA-LoRA's effectiveness arises from utilizing the the nonlinear capabilities of activation functions during early training. The 30\% decay schedule achieves the best performance by providing sufficient nonlinear capacity for initial feature adaptation alongside adequate linear training for convergence integration. Therefore, we use this 30\% decay setting by default in nearly all subsequent experiments (SFT in Section~\ref{subsec:sft_experiments} and GRPO in Section~\ref{subsec:rl_experiments}).

\section{Conclusion}
We propose Activation Function Annealing (AFA) as a simple yet effective way to enhance the expressive power of mergeable adapters like LoRA. By gradually changing non-linear activations to linear functions during training, AFA allows parameter-efficient fine-tuning to perform better without losing representational information for merging. Our experiments in supervised fine-tuning, reinforcement learning, and speculative decoding show that AFA-based adapters can reduce the performance gap with full-model adaptation. These gains come from small architectural changes and do not add any extra cost when deploying models for inference.

This work suggests new possibilities for designing efficient model adapters. In future research, it will be meaningful to explore adaptive annealing schedules for different tasks, apply this strategy to other other components in neural networks with diverse architectures, and automatically choose activation functions for specific domains. We believe that activation function annealing is a strong foundation for building high-performance yet practical adaptation techniques.

\section{Limitations}
\label{subsec:limitations}

While AFA-LoRA performed well across multiple tasks, this study still has some limitations that warrant further investigation in future work.

AFA-LoRA introduces additional hyperparameters, including annealing time ($T_{\text{start}}$, $T_{\text{end}}$), activation function type (ReLU, SiLU, GeLU), and their placement in the adapter. While experiments show that 30\% annealing time and SiLU activation perform well on most tasks, these optimal configurations may vary across different tasks, datasets, or model sizes, requiring task-specific tuning. Additionally, the continuous gradient variations during annealing may affect training stability, particularly in large-scale distributed training or resource-limited environments. Future work could explore adaptive annealing strategies, more stable optimization methods, and systematic evaluations under various hardware setups to enhance the method's robustness and practicality.

\bibliography{custom}

\nocite{schulman2017proximal}       
\nocite{yu2025dapo}                 
\nocite{sheng2024hybridflow}        
\nocite{sutton1998reinforcement}    
\nocite{loshchilov2017decoupled}    
\nocite{peft}                       
\nocite{agarap2018deep}             

\nocite{zheng2023judging}           
\nocite{cobbe2021gsm8k}             
\nocite{chen2021evaluating}         
\nocite{alpaca}                     




\end{document}